\documentclass[10pt,twocolumn,letterpaper]{article}

\usepackage{iccv}
\usepackage{times}
\usepackage{epsfig}
\usepackage{graphicx}
\usepackage{amsmath}
\usepackage{amssymb}

\usepackage[pagebackref,breaklinks,colorlinks]{hyperref}

\usepackage[capitalize]{cleveref}
\crefname{section}{Sec.}{Secs.}
\Crefname{section}{Section}{Sections}
\Crefname{table}{Table}{Tables}
\crefname{table}{Tab.}{Tabs.}

\iccvfinalcopy 

\DeclareMathOperator*{\argmax}{argmax}

\usepackage{color}
\definecolor{darkorange}{rgb}{1.0, 0.55, 0.0}

\newcommand{\w}{\mathbf{w}}

\newcommand{\x}{\mathbf{x}}
\newcommand{\xbar}{\mathbf{\bar{x}}}

\newcommand{\p}{\mathbf{p}}

\newcommand{\n}{\mathbf{n}}
\newcommand{\colr}{\mathbf{c}}
\newcommand{\view}{\mathbf{d}}

\newcommand{\mask}{\mathbf{M}}
\newcommand{\loss}{\mathcal{L}}
\newcommand{\renderer}{\Pi}

\newcommand{\skinf}{f_{\boldsymbol{\omega}}}

\newcommand{\albedof}    {a_{\boldsymbol{\omega'}}}
\newcommand{\rflectancef}{r_{\boldsymbol{\omega''}}}
\newcommand{\pose}{\boldsymbol{\theta}}
\newcommand{\posenorm}{\tilde{\boldsymbol{\theta}}}
\newcommand{\R}{\mathbb{R}}

\begin{document}


\title{VINECS: Video-based Neural Character Skinning}  

\author{ 
    Zhouyingcheng Liao\textsuperscript{1,2} \qquad 
    Vladislav Golyanik\textsuperscript{1} \qquad 
    Marc Habermann\textsuperscript{1} \qquad 
    Christian Theobalt\textsuperscript{1} \qquad 
    \\
    \quad
    \small{\textsuperscript{1}Max Planck Institute for Informatics, Saarland Informatics Campus} \qquad \small{\textsuperscript{2}The University of Hong Kong}
} 


\maketitle


\begin{abstract}
Rigging and skinning clothed human avatars is a challenging task and traditionally requires a lot of manual work and expertise. 
Recent methods addressing it either generalize across different characters or focus on capturing the dynamics of a single character observed under different pose configurations. 
However, the former methods typically predict solely static skinning weights, which perform poorly for highly articulated poses, and the latter ones  either require dense 3D character scans in different poses or cannot generate an explicit mesh with vertex correspondence over time. 
To address these challenges, we propose a fully automated approach for creating a fully rigged character with pose-dependent skinning weights, which can be solely learned from multi-view video. 
Therefore, we first acquire a rigged template, which is then statically skinned.
Next, a coordinate-based MLP learns a skinning weights field parameterized over the position in a canonical pose space and the respective pose.
Moreover, we introduce our pose- and view-dependent appearance field allowing us to differentiably render and supervise the posed mesh using multi-view imagery.
We show that our approach outperforms state-of-the-art while not relying on dense 4D scans. 
\end{abstract}


%
%
\vspace{-12pt}
\section{Introduction} \label{sec:intro}
%
%
%
\begin{figure}
 \centering
	\includegraphics[width=0.9\linewidth]{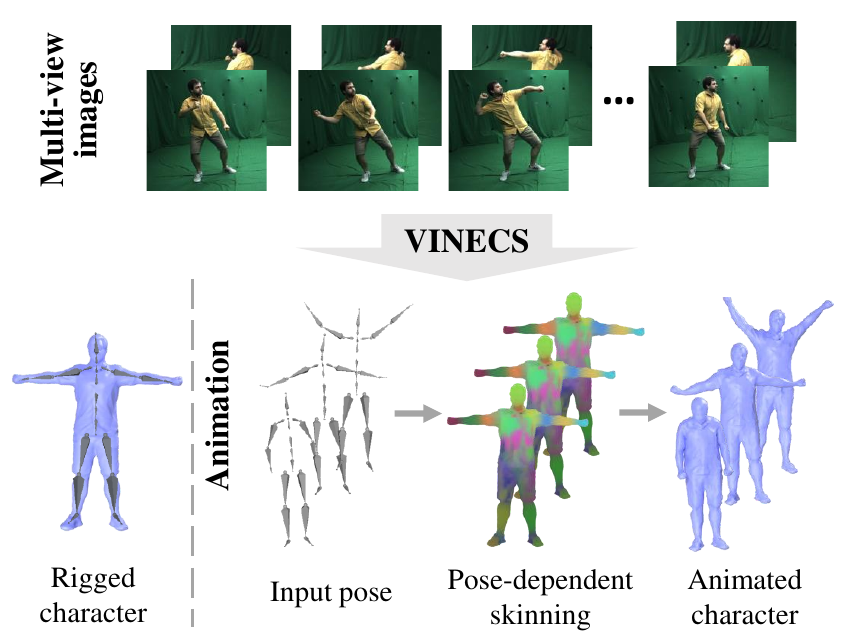}
	\vspace{-4pt}
	\caption
	{
	    \textbf{We propose the first end-to-end trainable method for generating a dense and rigged 3D character mesh including learned pose-dependent skinning weights solely from multi-view videos}.
        The results demonstrate that our model can generate visually-plausible geometry without requiring manual user input.
	}
	\label{fig:futureTeas}
	\vspace{-16pt}
\end{figure}
%
%
%
%
\par 
Animating 3D characters is a long-standing and challenging problem in computer graphics and vision. 
Traditionally, rigging, the process of positioning a sparse skeletal structure inside a dense 3D mesh, and skinning, the process of associating vertices with the skeleton, require large amounts of manual effort by experienced artists. 
This is not only expensive and time-consuming but also inherently difficult since one set of skinning weights is typically not ideal for all sorts of articulated character poses.
%
%
%
\par 
There are also methods for automated rigging and skinning.
Some recent works~\cite{neuroskinning, RigNet, skinningnet, liao2022pose} propose a learning-based solution for obtaining skinning and rigging given a single 3D character mesh.
However, they do not consider pose-dependent skinning correctives and lead to the typical appearance artifacts. 
Therefore, other techniques focus on pose-dependent skinning formulations accounting for pose-dependent variations and  deformations~\cite{Sloan01,rangescan, multiweight,  building, li2021learning}. 
Noteworthy, none of these methods tried to learn pose-dependent skinning weight correctives purely from \textit{2D image data}. 
Instead, they typically assume a 3D mesh is given, and most of them solely work on a fixed mesh resolution, which cannot be easily adjusted. 

%
%
%
\par 
To overcome these limitations, we propose VINECS, \textit{i.e.,} \textbf{VI}deo-based \textbf{NE}ural \textbf{C}haracter \textbf{S}kinning. 
Given solely a multi-view video of a human actor, our method creates a fully rigged, skinned, and textured 3D template (including hands) along with pose-dependent skinning weights while \textit{no manual editing} is required.
Moreover, VINECS enables multi-resolution character skinning since skinning weights can be sampled over a continuous 3D canonical space, which is not limited to any specific mesh resolution. 
\par 
For each character, we select one canonical pose frame, for which we extract a dense and textured 3D template mesh using implicit surface reconstruction techniques~\cite{wang2021neus}.
Given the template and the respective pose obtained from an off-the-shelf multi-view markerless capture system~\cite{captury}, we leverage an auto-rigging and skinning method, Pinocchio~\cite{baran07}, for obtaining an initial set of skinning weights. 
Note that Pinocchio produces only a \textit{static} set of skinning weights, which leads to the typical geometry artifacts when poses deviate too much from the canonical pose. 
Therefore, we propose a coordinate-based multi-layer perceptron (MLP)---which takes as input a 3D point on the surface of the template in canonical pose and the skeletal pose---and predicts the pose-dependent skinning weights at that point.
Since we want to learn the skinning weights solely from multi-view imagery, we design dedicated losses consisting of a silhouette loss and a
rendering loss. 
Specifically, we find that obtaining a well-behaved convergence is challenging only using a static texture for the rendering loss.  
Thus, we adopt the rendering formulation with an albedo network and a shadow network, which predict pose- and view-dependent appearance of our 3D character resulting in an improved convergence and skinning weights prediction.
%
%
%
In summary, our technical contributions are as follows:
\begin{itemize}
    \setlength{\parskip}{1pt} 
	\item An end-to-end trainable method for animation-ready explicit character creation, involving static template generation, rigging, and pose-dependent skinning solely using 2D multi-view videos. 
	\item A coordinate-based and pose-dependent skinning formulation, which enables multi-resolution skinning.
	\item A dedicated character appearance formulation and differentiable rendering enabling weakly supervised learning based on multi-view videos. 
\end{itemize}
We compare VINECS to the state-of-the-art and our results confirm that our method is a clear step toward accurate and automated creation of animatable 3D characters.
\par 
\textbf{Potential Impact.}
The focus of this work is \textit{not} to model deformable geometry~\cite{habermann19,habermann20,li2020deep,jiang2022hifecap} or 
photorealistic appearance~\cite{liu2021neural, peng2020neural, peng2021animatable, habermann21} of humans but on learning 3D characters with pose-dependent skinning from 2D observations only. 
We believe our method can greatly benefit many downstream tasks as many existing works leverage naked body models~\cite{loper15} for space canonicalization and feature assignment, fundamentally limiting them to clothing types of the same topology as the human body. 
As our rigging and skinning are accurate, easy-to-use, and robust to any type of clothing, our approach can stimulate research on the modeling of geometry and appearance of loose apparel. 
%

%
%
\section{Related Work} \label{sec:related}
\paragraph{Static Skinning.}
Linear Blend Skinning (LBS)~\cite{Magnenat-Thalmann1988:4, lbs} linearly blends rigid transformations of each bone to obtain a posed geometry.
However, this formulation can lead to a collapse of geometry near the skeleton joints, also called \textit{candy-wrappers effect}.
To overcome this, several follow-up works have been proposed using spherical representations~\cite{sphericalskinning}, sweep surfaces~\cite{Hyun2005SweepbasedHD}, curve-based skeleton parameterizations~\cite{Yang2006CurveSS}, Dual Quaternions~\cite{kavan07}, or by optimizing the centers of rotation~\cite{centersofrot}.
Other works focused more on computing the skinning weights itself by modeling the weight distribution as a heat diffusion process~\cite{baran07}, as illumination~\cite{boneglow}, or by voxelizing the mesh to increase robustness~\cite{geodesicbinding, voxelbinding}.
Later on, deformation editing was also modeled using a Laplacian energy formulation~\cite{boundedbiharmonic}, and user-defined splines on the mesh enable intuitive editing of skinning-based deformation behaviors~\cite{splineskinning}.
They all only assume a single geometry and a corresponding skeleton are given. 
%
%
%
\par 
In contrast, data-driven methods assume multiple instances of the same character under different poses or a large collection of rigged and skinned characters.
Earlier works~\cite{skinmeshani, robustaccurate} require meshes under different poses and then automatically detect skeleton joints and compute the skinning weights. 
Recently, NeuroSkinning~\cite{neuroskinning} was proposed as the first learning-based method predicting a static set of skinning weights given the mesh and a corresponding skeleton.
RigNet~\cite{RigNet} instead jointly predicts the skeleton as well as a set of skinning weights solely given a 3D mesh as input.
Therefore, they collected a large dataset of 3D characters including artist-designed skeletons and skinning weights.
SkinningNet~\cite{skinningnet} proposed a two-stream graph convolutional network architecture for improved skinning weight prediction given a mesh and the skeleton.
Yang et al.~\cite{yang2022object} propose to predict the 3D shape together with the rigging from a single image.
Liao et al.~\cite{liao2022pose} proposed a learning-based method which jointly learns to predict skinning weights and transfer poses.
%
%
%
\par 
What all the above methods have in common is that they do not explicitly account for dynamic or pose-dependent effects, such as muscle bulging or coarse cloth deformations.
In contrast, our proposed method learns pose-dependent skinning weights, which alleviate typical artifacts that arise when using a static set of skinning weights.
Moreover, it can also represent coarse deformations induced by the skeletal pose, such as coarse cloth deformations.
%
%

\textbf{Dynamic Methods.}
%
%
More closely related to our method are approaches that account for dynamic or pose-dependent effects, which are discussed next.
There are works~\cite{arcspline} that allow an artist to add pose-dependent effects by painting onto specialized textures or by adding so-called helper bones~\cite{bonehelpers}.
Alternatively, Kavan et al.~\cite{elasticity} proposed to find the optimal skinning weights by minimizing an elastic energy and adding joint-based deformers.
Interestingly, those methods do not assume any dynamic observations of the character while still modeling dynamic effects.
However, they either require manual editing~\cite{arcspline, bonehelpers} or can mostly model elasticity-based dynamics~\cite{elasticity}.
%
%
%
\par 
Similar to the static methods, several dynamic methods assume a (large) dataset of character meshes is given. 
Earlier works~\cite{rangescan, multiweight, building, capturingskin} account for dynamic effects by capturing a set of range scans depicting the subject in different poses.
Given a new pose, a set of pose-dependent blend weights is estimated, which linearly combines the set of example poses in the dataset or alternatively uses linear and radial basis functions~\cite{Sloan01}.
However, since the complexity of computation typically linearly scales with the number of example poses in the dataset, those methods cannot scale up to a very large number of examples (e.g., 20000 pose frames), which is in stark contrast to our approach that can deal with such an amount of data. 
There are also parametric naked human body models~\cite{anguelov05, correlated, loper15} learned by applying PCA over a large set of human scans, which can account for different shape- and pose-dependent deformations. 
In contrast, we build a highly-detailed rigged and skinned character that can wear (loose) clothing in a completely automated fashion. 
Li et al.~\cite{li2021learning} predict static skinning weights and pose-dependent neural blend shapes. 
They assume a dataset of dense meshes is provided, whereas we purely learn pose-dependent skinning weights from videos. 
SCANimate~\cite{scanimate} also requires dense point clouds, and learns a static set of skinning weights, while pose-dependent shapes are predicted by a pose-aware implicit field.
SNARF~\cite{chen2021snarf} represents the character as a pose-dependent neural implicit surface in the canonical pose space while the learned skinning weights are pose-agnostic.
Again, they require point cloud data while our method can be supervised directly on the video.
In summary, none of the above methods that consider pose-dependent effects investigated learning solely from 2D imagery. 
Thus, we are the first method demonstrating that dynamic effects in the form of pose-dependent skinning weights can be learned only from videos.
%
%
\par 
Recently, some works were proposed to skin and rig implicit surface representations. 
ARAH~\cite{arah} learns an implicit human model using neural rendering. 
However, as it relies on a SMPL template and an inverse root finding to convert from view space to canonical space, it easily fails when the garment deviates much from the body. 
Importantly, it also does not have the vertex correspondence over time.
TAVA~\cite{li2022tava} avoids these problems by removing the template assumption, which, however, leads to poor geometry recovery mainly due to their density-based scene representation.
%
%
%
\section{Method} \label{sec:method}
%
%
%
\begin{figure*}
	\centering
	\includegraphics[width=\linewidth]{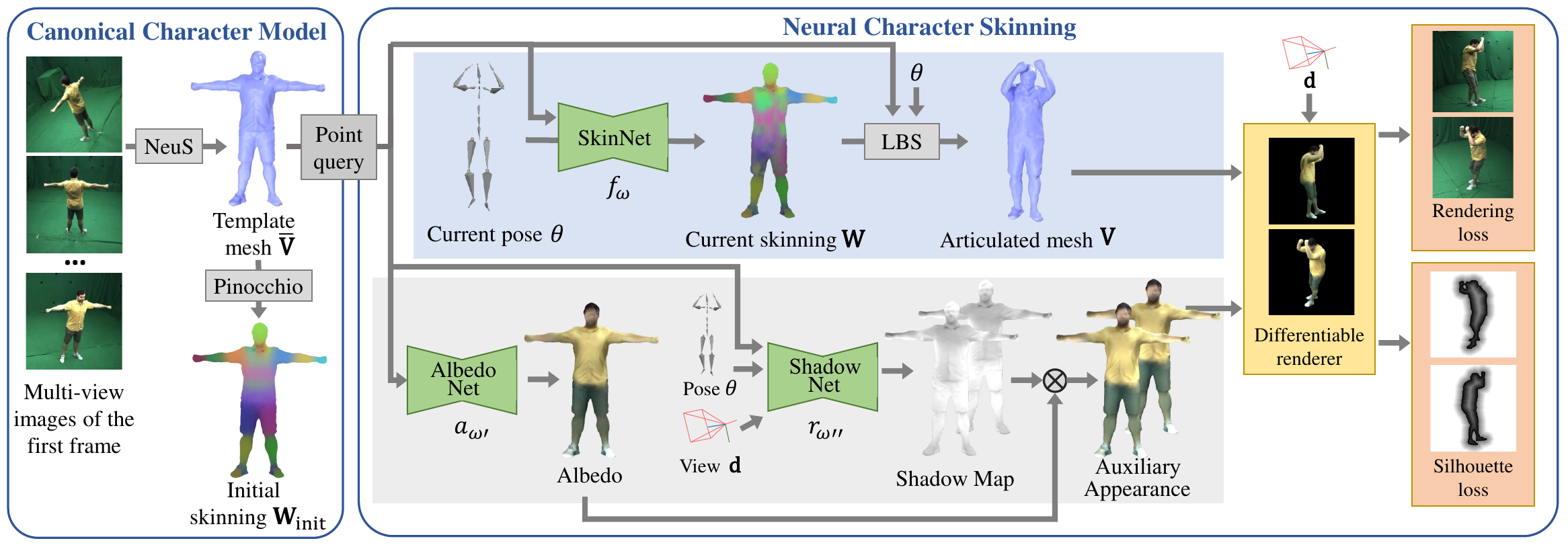}
	\vspace{-12pt}
	\caption
	{
	    \textbf{Method overview.}
	    Our method, VINECS, is a fully automated solution to template generation, rigging, and pose-dependent skinning solely using multi-view videos.
	    We first recover the skeletal poses using markerless motion capture~\cite{captury} and a static template mesh using implicit surface reconstruction~\cite{wang2021neus}.
	    An initial skinning is obtained by a heat diffusion process~\cite{baran07}.
	    Since one fixed set of skinning weights can lead to artifacts when significantly changing the character pose, we learn pose-dependent skinning weights using a coordinate-based MLP.
            An auxiliary appearance field is applied to better supervise the learning of pose-dependent skinning.
}
	\label{fig:overview}
	\vspace{-12pt}
\end{figure*}
%
%
%
Given synchronized and calibrated multi-view RGB videos of a human character in motion, our goal is to learn an animatable 3D human model with dynamic skinning, without any manual process, such as meshing, rigging, and skinning. 
To this end, we first extract the static geometry of the actor from one canonical pose frame (typically depicting the actor in a T-pose) of the recording and automatically compute initial skinning weights (Sec.~\ref{subsec:template}). 
However, the initial skinning weights are calculated only based on the static geometry of the human character, and, as studied in earlier works, one set of weights might be ideal for one pose while it can lead to strong artifacts in other poses. 
Thus, we propose a skinning network, called SkinNet, to predict the dynamic skinning weights as a function of the skeletal pose and 3D coordinates in the canonical space (Sec.~\ref{subsec:skinnet}).
Given the skeletal pose, the static geometry, and the pose-dependent skinning weights, we leverage Linear Blend Skinning (LBS) to obtain the posed geometry. 
We then use a differentiable renderer to weakly supervise the skinning weights solely based on multi-view imagery. 
Thereafter, we propose a novel appearance field (Sec.~\ref{subsec:texture}), which is composed of an albedo component and a pose- and view-dependent shadow/shading component.
In addition, we propose several regularization losses ensuring that the skinning weights are also robust to out-of-distribution poses and that the final animated geometry looks visually plausible (Sec.~\ref{subsec:loss}).
An overview of our method is illustrated in Fig.~\ref{fig:overview}.
Before we explain our method in more detail, we first discuss our data assumptions and introduce notations.
%
%
\par \textbf{Data Preparation.}
%
For a human subject, we record a multi-view video of the performance using $C$ calibrated and synchronized cameras with a total length of $F$ frames. 
For all frames $\mathcal{I}_{c,f}$, we calculate the foreground segmentation using color keying or background matting~\cite{BMSengupta20} and compute respective distance transform images $\mathcal{D}_{c,f}$.
Here, $c$ and $f$ refer to the respective camera and frame index.
We capture the size of the human skeleton and the motion using markerless motion capture~\cite{captury}.
The skeleton motion is parameterized by a root rotation $\boldsymbol{\alpha}_{f} \in \mathbb{R}^3$, a root translation $\mathbf{t}_{f} \in \mathbb{R}^3$, and the joint angles $\boldsymbol{\rho}_{f} \in \mathbb{R}^D$.
The full pose vector is defined as $\boldsymbol{\theta}_{f} = [\boldsymbol{\alpha}_{f},\mathbf{t}_{f},\boldsymbol{\rho}_{f}]^T  \in \mathbb{R}^{3+3+D}$.
%
\par \textbf{Notation.}
%
We denote vectors by bold lower-case letters, matrices by bold upper-case letters, and scalars and functions by lower-case letters. 
To represent a variable in the canonical space, we use a bar over the letter, i.e., $\xbar$.
For any point in canonical and global space, we use $\xbar$ and $\mathbf{x}$, respectively, and mesh vertices are denoted as $\bar{\mathbf{v}}_i$ and $\mathbf{v}_i$ where $i$ is the $i$th vertex.
Without loss of generality, we assume the frame $f$ is fixed in the following (if not explicitly mentioned otherwise) and omit the subscript for better readability.
%
%
\subsection{Canonical Character Model} \label{subsec:template}
As the first step, we acquire a canonical model of the actor.
Recently, implicit surface and coordinate-based representations~\cite{Park_2019_CVPR} have gained considerable attention due to their flexibility and comparably simple integration into learning frameworks, and they were also leveraged in the context of pose-dependent human deformation modeling~\cite{chen2021snarf}.
Moreover, we compared the reconstruction accuracy on the first frame $f=0$ between classical multi-view stereo and implicit reconstruction method called NeuS~\cite{wang2021neus} in the supplemental material.
Interestingly, NeuS has a much higher level of detail and comparably less noise on the surface. 
Thus, we use NeuS~\cite{wang2021neus} to reconstruct the mesh in the first frame, also called the canonical frame/space. 
%
\par \textbf{Implicit vs. Explicit Canonical Model.}
%
In stark contrast to Chen et al.~\cite{chen2021snarf}, we run Marching Cubes~\cite{marchingcubes} to convert the implicit surface into an explicit mesh and obtain per-vertex colors from NeuS by setting the viewing direction to the normal direction of the mesh.
The reason for choosing an explicit mesh in canonical space is two-fold: 
1.) It avoids running the rather slow Marching Cubes algorithm for every potentially posed mesh.
2.) We avoid the backward skinning~\cite{chen2021snarf}, which boils down to a correspondence search that has to be solved using iterative optimization.
3.) We can maintain correspondence on the surface across poses, which is especially useful when texturing and other editing have to be applied.
Thus, we aim for a hybrid strategy for template acquisition, i.e., using the state-of-the-art method for implicit surface reconstruction and then converting it to an explicit mesh.
%
\par \textbf{Meshing and Initial Skinning.}
%
During training, we downsample the template mesh to $N$ vertices where $\bar{\mathbf{v}}_i \in \mathbb{R}^3$ denotes the $i$th vertex.
Given the template mesh and the corresponding posed skeleton, i.e. $\boldsymbol{\theta}_{f}$, we utilize Pinocchio~\cite{baran07} to obtain an initial set of skinning weights $\w_{\mathrm{init},i} \in \R^J$ for each vertex $i$ (see Fig.~\ref{fig:qualitative2}).
Here, $J$ denotes the number of skinning joints.
Given these weights and a skeletal pose $\pose$, we can transform the vertex $\bar{\mathbf{v}}_i$ from the canonical space to the posed space using LBS~\cite{lbs}:
%
\begin{equation} \label{eq:lbs}
    \mathbf{v}_i = LBS(\bar{\mathbf{v}}_i, \w_{\mathrm{init},i}, \pose) : \R^3 \times \R^J \times \R^D \rightarrow \R^3.
\end{equation}
%
\par \textbf{Mesh Parsing.}
%
Since different body parts and materials (skin vs. clothing) have different deformation behaviors, we compute per-vertex human parsing labels.
We render the 3D model into all camera views and apply a 2D human parsing approach~\cite{li2020self} on each view. 
We then perform max-voting to obtain the per-vertex human parsing labels.
More details can be found in the supplementary material. 
Note that so far, we obtained a segmented, rigged, and statically skinned character model \textit{without} any manual user input.
%
%
\subsection{Pose-dependent Skinning Field} \label{subsec:skinnet}
So far, the initial skinning weights obtained in Sec.~\ref{subsec:template} have two main limitations:
1.) They are solely computed on a single static pose leading to artifacts such as candy wrappers and mesh distortions if the new pose significantly differs from the canonical one.
2.) The set of weights is computed per-vertex.
Thus, the resolution of the mesh has to be fixed beforehand and any change in the mesh requires a new computation of skinning weights or some complicated weight transfer from one mesh to the other.
%
\par
%
To resolve this, we propose SkinNet $\skinf$, which is a coordinate-based MLP where $\boldsymbol{\omega}$ denotes the trainable network weights. 
Given any 3D point $\xbar$ in the canonical space, $\skinf$ predicts its skinning weight $\w $ conditioned on the normalized human pose $\posenorm$, which can be formulated as:
%
\begin{equation} \label{eq:skin_net}
    \w_{\posenorm} = \skinf(\xbar, \posenorm) : \R^3 \times \R^D \rightarrow \R^J,
\end{equation}
%
where $\xbar$ and $\posenorm$ are concatenated before being fed into the MLP.
Here, the normalized pose $\posenorm$ is obtained by neglecting the global translation as well as the yaw angle, i.e., the rotation around the spine.
The intuition behind is that the skinning weights do not depend on where in global 3D space the person performs a respective pose, and, similarly, the skinning weight prediction should be agnostic to which direction the person is facing. 
Thus, we mask those degrees of freedom before feeding the pose vector into the network.
To obtain the transformed point $\mathbf{x}$ in posed space, one can insert Eq.~\eqref{eq:skin_net} into Eq.~\eqref{eq:lbs} resulting in:
%
\begin{equation} \label{eq:final_skin}
    \mathbf{x} = LBS(\xbar, \skinf(\xbar, \posenorm), \pose).
\end{equation}
%
Revisiting the drawbacks of initial skinning discussed above, the SkinNet now supports pose-dependent skinning weights such that the artifacts of static skinning can be reduced.
Moreover, since SkinNet is a coordinate-based MLP, it supports arbitrary mesh resolution since each vertex is queried independently, which is in stark contrast to an architecture that outputs all skinning weights at once and where the final tensor shape is fixed to a specific mesh resolution (see the supplemental document, which evaluates this design choice).
Details about the architecture are provided in the supplemental document as well.
Next, we explain how we learn a neural dynamic appearance model of the actor, which can then be leveraged in our image-based losses.
%
%
\subsection{Auxiliary Appearance Field} \label{subsec:texture}
We can now pose the geometry with our pose-dependent skinning weight formulation.
However, since we want to supervise SkinNet solely on multi-view images, we also have to model the appearance of the actor.
A naive choice would be to just leverage the static vertex colors acquired from NeuS. 
However, we found that this texture contains baked-in shadows and wrinkles, and cannot account for the pose- and view-dependent appearance changes harming the training of the SkinNet. 
We also refer to our ablation studies in Sec.~\ref{sec:ablation}.
To address this issue, we propose an auxiliary appearance field, which consists of an albedo field predicting a pose- and view-\textit{agnostic} color for a given point $\xbar$ in the canonical space and a shadow field predicting the pose- and view-\textit{dependent} shadow/shading properties of $\xbar$. 
Important to note here is that we are \textit{not} interested in creating a highly photorealistic appearance model of the human, but we mainly use it as an auxiliary tool to better supervise the learning of the skinning weight network.
%
\par
More specifically, the albedo field, referred to as \textit{AlbedoNet}, is an MLP that
predicts a static albedo value $\albedof(\xbar)=\mathbf{a} \in \mathbb{R}^3$ for a given 3D sample point in the canonical space.
Here, $\boldsymbol{\omega'}$ are the trainable weights of the albedo network.
Moreover, the shadow field is also parameterized by an MLP $\rflectancef(\xbar, \pose, \n, \view) = s \in \mathbb{R}^+$, called \textit{ShadowNet}, which predicts a scalar multiplier accounting for color changes due to shadows and shading.
Note that we assume that such effects happen uniformly across all color channels and, thus, we are only predicting a single scalar instead of a scaling factor per color channel.
In addition to taking the point in canonical space, the shadow field also takes the skeletal pose $\pose$ as input as well as the surface normal $\n$ of $\mathbf{x}$ in global space and the viewing direction $\view$, which allows to potentially model pose- and view-dependent lighting effects such as shadows and shading.
$\boldsymbol{\omega''}$ denotes the respective network weights.
Details about the architectures are provided in the supplemental document.
The final color $\colr \in \mathbb{R}^3$ for point $\mathbf{x}$ is computed as 
%
\begin{equation} \label{eq:final_color}
    \colr = \albedof(\xbar)\rflectancef(\xbar, \pose, \n, \view).
\end{equation}
%
\subsection{Multi-view Video-based Supervision} \label{subsec:loss}
Next, we describe our dedicated supervision strategy that solely requires multi-view imagery.
We first introduce the individual loss terms and then describe our training strategy.
Importantly, we can obtain a posed template mesh $\mathbf{V} \in \mathbb{R}^{N \times 3}$ and the pose- and view-dependent per-vertex colors $\mathbf{C}_c \in \mathbb{R}^{N \times 3}$ by evaluating Eq.~\eqref{eq:final_skin} and Eq.~\eqref{eq:final_color} for all vertices $\bar{\mathbf{v}}_i$ and views $c$.
Moreover, the following losses that are applied to the deformed geometry and its appearance can directly backpropagate into the respective networks since our formulation is fully differentiable.
%
%
\par \textbf{Silhouette Loss.}
This loss~\cite{habermann21} ensures that the projected posed geometry projects onto the foreground masks for all views (see supp. material).
Since our pose-dependent skinning field Eq.~\eqref{eq:final_skin} is differentiable with respect to the SkinNet weights $\boldsymbol{\omega}$, the silhouette loss can provide supervision and learn pose-dependent skinning by matching the posed model silhouette and the image silhouettes.
%
%
\par \textbf{Rendering Loss.}
Although the silhouette loss ensures that the posed geometry matches the multi-view silhouettes, it cannot supervise drifts on the visual hull.
Supervision directly from the RGB images can help with resolving this issue since appearance features can constrain drifts in the image plane.
Thus, we introduce the rendering loss
%
\begin{equation}
    \loss_\mathrm{rend} = \sum_{c=1}^C \lVert \renderer_c (\mathbf{V}, \mathbf{C}_c) - \mathcal{I}_c \rVert_1,
\end{equation}
%
which leverages a differentiable renderer $\renderer_c$ based on the one of Habermann et al.~\cite{habermann21} to render the posed and colored model into all camera views $c$ that is then compared to the ground truth image $\mathcal{I}_c$. 
Importantly, this loss can supervise both the pose-dependent skinning field as well as the pose- and view-dependent appearance field.
%
%
\par \textbf{Laplacian Loss.}
Since the data terms alone can still lead to degenerate results, we also introduce some regularization on the geometry and the skinning weights.
First, we regularize the posed geometry with a Laplacian loss
%
\begin{equation}
    \loss_\mathrm{lap} = \sum_{i=1}^N \parallel \mathbf{v}_i - \frac{1}{|\mathcal{N}_i|} \sum_{k \in \mathcal{N}_i} \mathbf{v}_k \parallel_2^2
,
\end{equation}
%
where $\mathcal{N}_i$ is the indices of one-ring neighborhood of vertex $i$.
This regularizer ensures locally smooth geometry.
%
%
\par \textbf{Skinning Regularization.}
We found that SkinNet can overfit the training poses and respective deformations by predicting non-local skinning weights, e.g., the foot joint has a non-zero skinning weight for vertices on the shoulder. 
This, however, harms generalization to novel poses at test time.
To prevent this, we regularize the skinning weights by constraining their values based on the geodesic distance $d_\mathrm{geo}$ between the vertex and the initial body part it belongs to.
This can be formalized with the following loss
%
\begin{equation}
    \loss_\mathrm{skin} = \sum_{i=1}^N \sum_{j=1}^J 
    \w_{i,j}  
    \left( 
    \frac{ \min_{k \in \mathcal{A}_j}d_\mathrm{geo}(\bar{\mathbf{v}}_i, \bar{\mathbf{v}}_k)}
    {d_\mathrm{geomax}} 
    \right) ^r ,
\end{equation}
\begin{equation}
 \mathcal{A}_j = \{ k | j = \argmax_{t} \w_{\mathrm{init},k,t} \}, 
\end{equation}
%
where $\mathcal{A}_j$ is the set of vertices, which have the highest skinning weight with respect to joint $j$. $r=3$ is a hyper-parameter which controls how penalization increases as the geodesic distance increases.
$d_\mathrm{geomax}$ is the maximum geodesic distance between any of the vertex pairs.
%
%
\par \textbf{Part-wise Regularization.}
In contrast to the clothed areas, the skin part has rather static skinning. 
Thus, we use the initial skinning to regularize the predicted skinning weights of the skin part, i.e., the set $\mathcal{G}$ of vertices, which have the parsing label \textit{skin}. 
However, even for skin vertices, we found that the initial skinning weights around the joints are not accurate while the rigid parts (i.e., the vertices away from joints) usually have correct initial skinning. 
We identify the set of rigid vertices as $\mathcal{R} = \{ k | \max (\mathbf{w}_{\mathrm{init},k}) > u \}$ where $u=0.95$.
Thus, our part loss
%
\begin{align}
    \loss_\mathrm{part} = \sum_{i \in \mathcal{R} \cap \mathcal{G} } \parallel \skinf(\bar{\mathbf{v}}_i, \posenorm) - \mathbf{w}_{\mathrm{init},i} \parallel_2^2
    ,
\end{align}
%
enforces that the predicted skinning weights are close to the initial ones \textit{iff} the human parsing label of a vertex is the skin ($\mathcal{G}$) and the vertex belongs to the rigid part ($\mathcal{R}$).
%
%
\par \textbf{Training Strategy.}
Our training proceeds in four stages. 
First, we train SkinNet \textit{without} using the rendering loss $\loss_\mathrm{rend}$.
This step is required to ensure that the posed mesh roughly overlaps with the foreground masks.
Next, the AlbedoNet is trained using only the rendering loss $\loss_\mathrm{rend}$ while the SkinNet weights $\boldsymbol{\omega}$ are fixed. 
Importantly, the AlbedoNet is \textit{not} pose-dependent, but we optimize the weights $\boldsymbol{\omega}'$ across all frames.
Afterward, the AlbedoNet and the SkinNet are fixed and we train the ShadowNet using the rendering loss $\loss_\mathrm{rend}$.
Again we train on all training frames.
Lastly, since we now have a pose- and view-dependent appearance model available, we refine the SkinNet weights including the rendering loss.
We found this gives the best performance, as validated by our results.
%
%
\section{Results} \label{sec:results}
We now qualitatively and quantitatively evaluate our approach, compare it to previous works, and perform ablation studies evaluating our proposed components in more detail.
%
%
\par 
\textbf{Dataset.}
We evaluate our method on three subjects wearing tight and loose clothing (pants or skirts). 
Two of them are from the DynaCap dataset~\cite{habermann21} and they are referred as \textit{D2} and \textit{D5}. 
\textit{D2} is wearing short pants and a T-shirt while \textit{D5} is in a skirt.
The number of cameras varies between $94$ and $101$ for those two sequences and it consists of around 19000 training frames per subject depicting the actor in various articulated poses.
For evaluation, there is a separate testing set of around 7000 frames per subject.
For convenience, we keep 1 out of 10 frames for training and evaluation.
Additionally, we captured a new subject, \textit{V6}, wearing a T-shirt and trousers using 116 cameras.
We captured a separate training (17730 frames) and testing sequence (5000 frames).
Importantly, for this sequence, we recover hand gestures and a respective 3D model including hands.
%
%
\par 
\textbf{Metrics.}
For evaluation, we reconstruct the dense 3D geometry for the testing frames using multi-view stereo reconstruction~\cite{agisoftphotoscan}.
We do not use NeuS here as it is very time-consuming.
Those dense meshes serve as ground-truth scans and we compare the posed meshes recovered by our method as well as competing methods in terms of the Chamfer distance and the Hausdorff distance going from the recovered mesh to the ground-truth scan (M2S) and vice versa (S2M). 
We report the average results over all testing frames.
All metrics are reported on centimeter scale.
%
%
\par 
\textbf{Competing Methods.}
We compare to SCANimate~\cite{scanimate} and SNARF~\cite{chen2021snarf}, which, in stark contrast to our work, require dense point clouds for training. 
They learn a static set of skinning weights and a pose-aware implicit shape.
Since our approach mostly focuses on skinning, we compare to SCANimate with and without pose-aware shape, referred to as SCANimate and SCANimate* in the following.
Finally, we compare to RigNet~\cite{RigNet}, which can jointly predict a rigging skeleton and static skinning weights solely given the template.
Unfortunately, SkinningNet~\cite{skinningnet} and NeuroSkinning~\cite{neuroskinning} do not provide their code and, thus, we cannot compare to them.
For more details and comparisons to other (slightly less) related works~\cite{li2022tava,arah}, we refer to the supplemental document.

%

%
\subsection{Qualitative Results} \label{sec:qualitative}
%
%
%
\begin{figure*}
	\includegraphics[width=\linewidth]{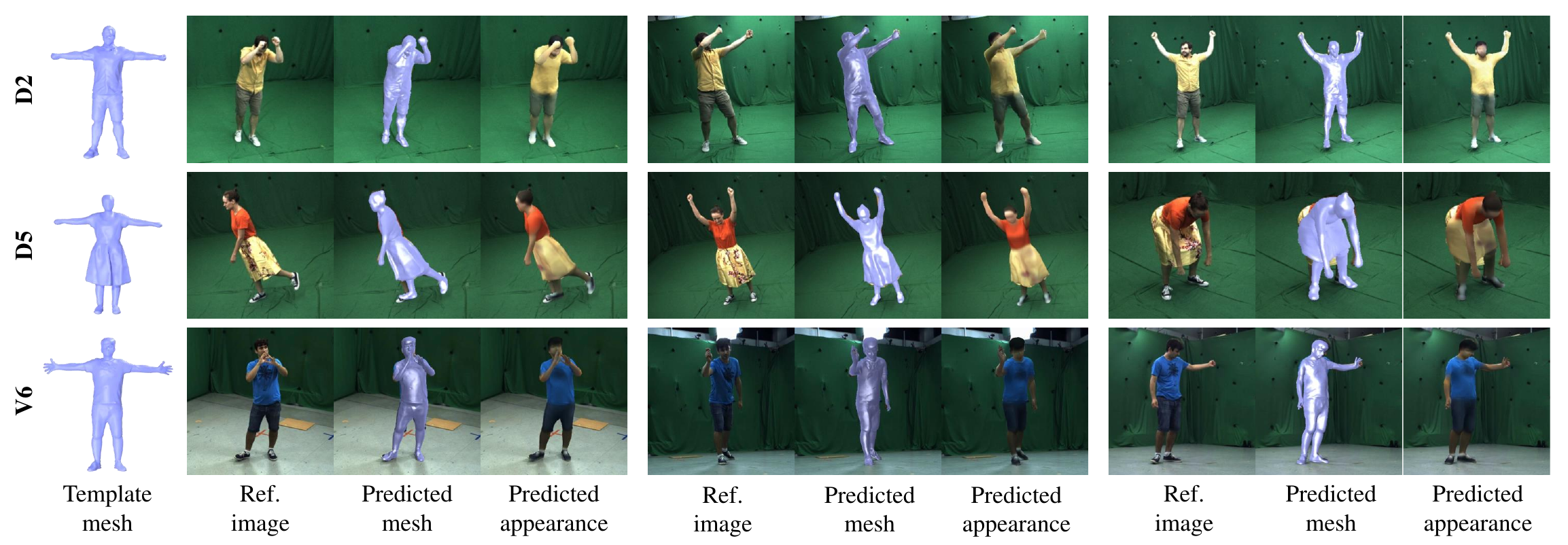}
	\vspace{-10pt}
	\caption
	{
	    Qualitative results. 
	    From left to right:
	    Recovered static character template.
	    Reference image showing a test pose.
	    Our posed character overlayed onto the reference image.
	    The precise overlay of the posed geometry confirms the effectiveness of our approach.
	}
	\label{fig:qualitative}
 \vspace{-10pt}
\end{figure*}
%
%
%
In Fig.~\ref{fig:qualitative}, we qualitatively evaluate our approach. 
In the first column, we show the reconstructed static geometry for different subjects. 
In the following columns, we show the characters in novel poses from the test set overlayed onto the respective images. 
Note that our posed character nicely matches the reference image, which confirms that our pose-dependent skinning indeed generalizes to novel poses and results in accurately posed meshes.
%
%
%
\begin{figure}
	\includegraphics[width=0.9\linewidth]{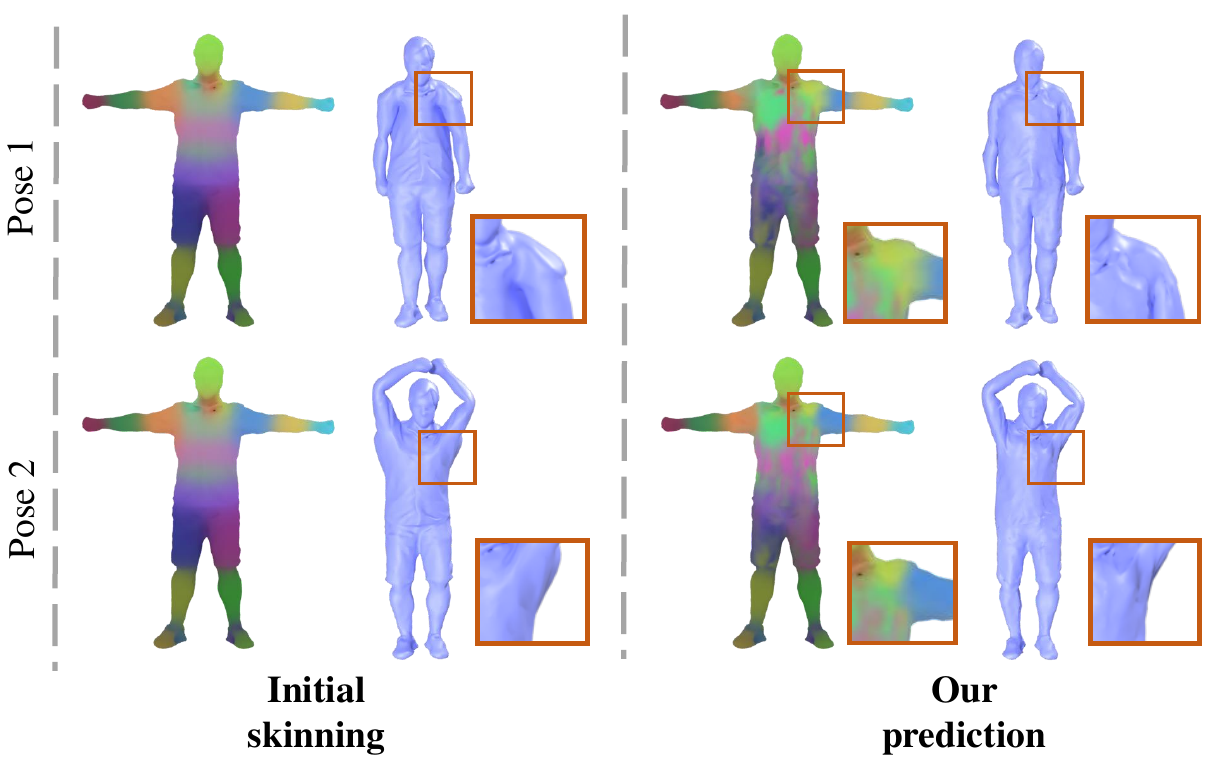}
	%
	\caption
	{
	    Qualitative results showing the pose-dependent skinning weights. 
	    As a reference, we show the result using the initial static skinning weights. 
	    Notably, using static skinning weights leads to artifacts and worse deformation.
	    When using our pose-dependent skinning, one can see a clear improvement in the final result and we also show how the weights change over different poses.
	}
	\label{fig:qualitative2}
		\vspace{-12pt}
\end{figure}
%
%
%
In Fig.~\ref{fig:qualitative2}, we further visualize the effect of the pose-dependent skinning weights obtained by our approach.
The initial skinning weights are fixed across poses leading to the typical skinning artifacts and wrong deformation behavior.
In contrast, our proposed pose-dependent skinning weights adapt according to the pose and as a result the deformations look more natural.
%
%
\subsection{Comparison} \label{sec:comparison}
%
%
%
%
\begin{table*}
\begin{center}
\begin{tabular}{|c|c|c|c|c|c|c|c|c|c|}
    \hline
    \multicolumn{10}{|c|}{\textbf{\textit{Quantitative Geometry Comparison}}} \\
    \hline
    \textbf{Subject} & \multicolumn{3}{|c|}{D2} & \multicolumn{3}{|c|}{D5} & \multicolumn{3}{|c|}{V6} \\
    \hline
    \textbf{Method}  & \textbf{Chamfer}$\downarrow$  & \textbf{M2S}$\downarrow$ & \textbf{S2M}$\downarrow$ 
    & \textbf{Chamfer}$\downarrow$  & \textbf{M2S}$\downarrow$ & \textbf{S2M}$\downarrow$
    & \textbf{Chamfer}$\downarrow$  & \textbf{M2S}$\downarrow$ & \textbf{S2M}$\downarrow$ \\
    \hline
    Pinnochio~\cite{baran07} 	                             &  3.760               &  2.162         &   1.599    &  5.077               &  2.811 &   2.267 &  3.358                &  1.871 &   1.487  \\
    SCANimate$^*$~\cite{scanimate} 	                            &   3.750               &  2.099         &   1.650  &   5.453               &  2.965         &   2.488   &   3.502               &  1.890         &   1.612   \\
    RigNet~\cite{RigNet}                                  &   3.599               &  2.078         &   1.521    &   4.989               &  2.836         &   2.153  &   3.369               &  1.894         &   1.475  \\
    \textbf{Ours} 	                        &   \textbf{3.034}               &  \textbf{1.746}         &   \textbf{1.288}     &   \textbf{4.512}               &  \textbf{2.442}         &   \textbf{2.070}   &   \textbf{2.993}               &  \textbf{1.719}         &   \textbf{1.274}    \\
    \hline
     SCANimate~\cite{scanimate} 	                            &   2.842               &  1.646         &   1.196  &   4.982               &  2.284         &   2.698    &   3.154               &  1.714         &   1.440      \\
    SNARF~\cite{chen2021snarf} &   3.306   &   1.981   &   1.325   &   4.560   &   2.405   &   2.154   &   3.489   &   1.952   &   1.537 \\
    \hline
    \end{tabular}
    \end{center}
    \vspace{-6pt}
    \caption
    {
    Comparison to previous work.
    Note that our method obtains better results than other skinning-based approaches.
    It is even close to SCANimate and SNARF, which require dense pointclouds while we solely use multi-view video.
    Importantly, they focus on learning pose-aware shapes while our goal is to obtain a fully rigged and skinned character in a completely automated fashion. 
    }
	\label{tab:quantitative}	
\end{table*}
%
%

%
In Tab.~\ref{tab:quantitative}, we compare our results to the previous works.
Note that we achieve the most accurate results when compared to works only focusing on character skinning. 
This validates the necessity of pose-dependent skinning weights since other works rely on static ones. 
In this table, SCANimate and SNARF predict a pose-aware shape while assuming dense point clouds are given. 
Since it is not the scope of this work to predict pose-dependent non-rigid deformations and our method solely learns from video data, we placed these works separately.
Nonetheless, it is worth noting that our approach performs better than SNARF for all subjects, and is close to SCANimate in terms of accuracy for subject \textit{D2} and \textit{V6} even though our method does not require point clouds for training.
Moreover, we show improved accuracy compared to SCANimate for subject \textit{D5}, who wears loose clothing. 
We found that SCANimate performs worse on subjects with loose clothing as they rely on SMPL while our method does not have such a restriction.
\par 
%
%
%
\begin{figure*}
	\centering
	\includegraphics[width=0.95\linewidth]{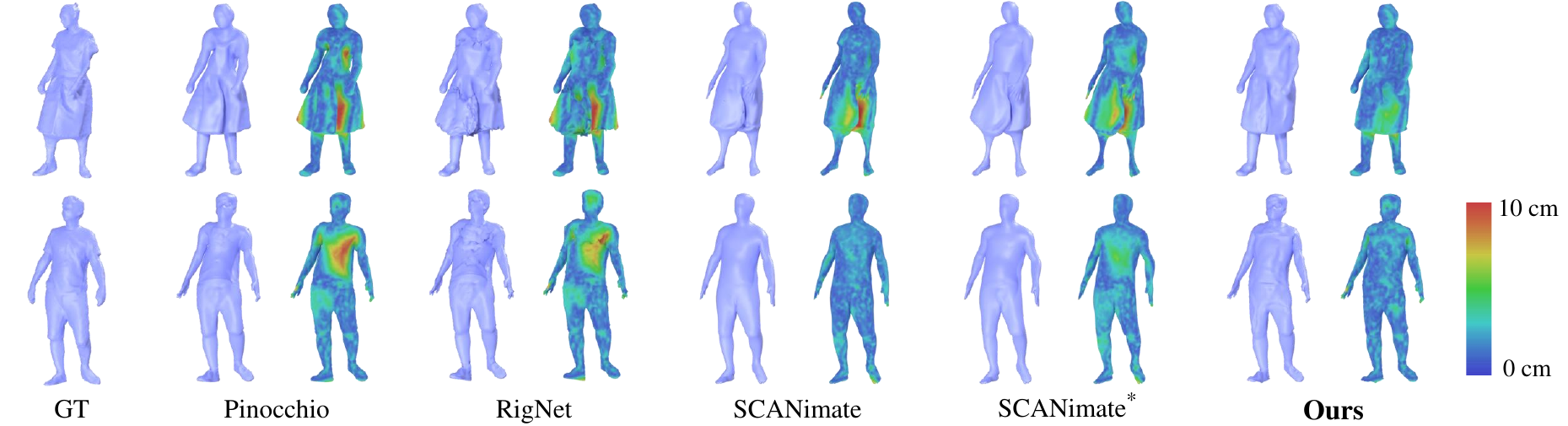}
	\caption
	{
	    Qualitative comparison.
	    For each method, we visualize the recovered posed geometry as well as the per-vertex error map when comparing the ground truth in terms of M2S.
	    Note that our method consistently shows the lowest error and also has the least visual artifacts.
	}
	\label{fig:comparison}
	\vspace{-8pt}
\end{figure*}
%
%
%
We also visually compare to those works in Fig.~\ref{fig:comparison}.
We can see that also qualitatively our method achieves a lower error especially in the regions of the joints where typically most of the skinning artifacts become visible.
This can be nicely seen in the error maps, which visualize the per-vertex error. 
Further, the posed meshes clearly show fewer artifacts for our approach compared to previous work.
Note that the entire character including the geometry, rigged skeleton, as well as the pose-dependent skinning can be obtained solely from multi-view video without any manual user input.
%
%
\subsection{Ablation} \label{sec:ablation}
%
%
%
\begin{table}
\begin{center}
\begin{tabular}{|c|c|c|c|}
    \hline
    \multicolumn{4}{|c|}{\textbf{\textit{Ablation Study on D2}}} \\
    \hline
    \textbf{Method}  & \textbf{Chamfer}$\downarrow$  & \textbf{M2S}$\downarrow$ & \textbf{S2M}$\downarrow$ \\
    Initial weights 	                                    &   3.761               &  2.162 &   1.599         \\
    Static weights 	                                         &   3.354               & 1.935          &   1.418      \\ 
    w/o $\mathcal{L}_\mathrm{rend}$ 	                                       &    3.116               &    1.783      &   1.333        \\ 
    w/ albedo-only	                                        &   3.047               &  1.750         &   1.297        \\ 
    w/ NeuS appearance 	                                  &   3.152               &  1.786         &   1.365           \\ 
    w/ NeuS albedo 	                                  &   3.137              & 1.795         &  1.342          \\

    w/ displacements	                                  &  3.191              & 1.805          &   1.386            \\ 
    \hline
    Ours (10 views)	                                  &   3.129              &  1.791          &  1.337            \\ 
    \textbf{Ours} 	                        &   \textbf{3.034}               &  \textbf{1.746}         &   \textbf{1.288}        \\  
    \hline
    \end{tabular}
    \end{center}
    \vspace{-6pt}
    \caption
    {
    We compare our proposed formulation against several baselines. 
    Note that all design choices gradually improve the overall accuracy confirming the effectiveness of our VINECS. 
    }
	\label{tab:ablation}
\end{table}
%
%

%
Next, we evaluate our design choices in Tab.~\ref{tab:ablation}.
First, we evaluate the effect of pose-dependent skinning weights.
Therefore, we compare our final result to the initial skinning, which we obtain by leveraging Pinocchio~\cite{baran07} and a version of our method where we use our supervision scheme, but optimize a static set of skinning weights, i.e., the weights are fixed across all poses.
From the results, we can see that pose-dependent weights significantly improve the accuracy over the two baselines 
proving the effectiveness of our formulation. 
%
%
%
\begin{figure}
	\includegraphics[width=\linewidth]{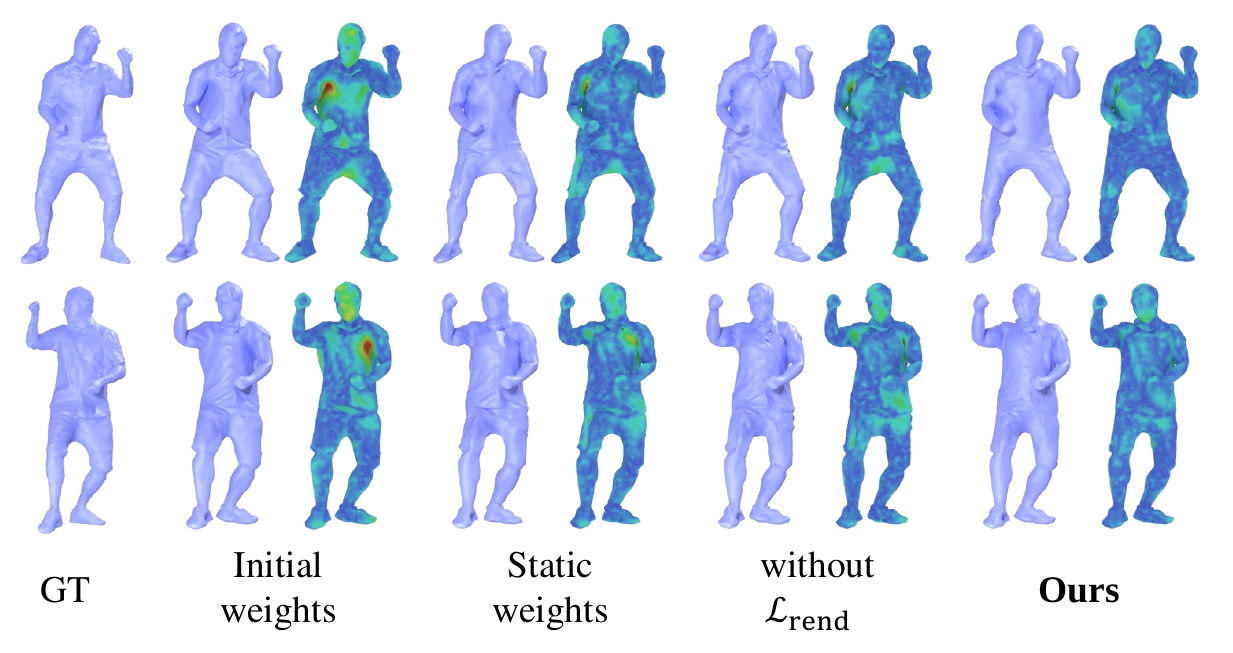}
	\vspace{-12pt}
	\caption
	{
	    We compare the initial skinning, the static skinning optimized using our supervision scheme, ablating the rendering loss, and our pose-dependent skinning. 
	    We outperform all baselines.
	}
	\label{fig:ablation}
		\vspace{-12pt}
\end{figure}
%
%
%
This is also visualized in Fig.~\ref{fig:ablation}. 
\par 
Another key aspect when learning those weights solely from video data is our proposed supervision scheme.
When we do not employ the dense rendering loss or soley use the albedo without shadow estimation, the performance drops. 
Moreover, we also evaluate the accuracy when using a dense rendering loss with the static appearance from NeuS and using NeuS appearance as albedo while the shadow network is learned.
We found that our proposed rendering scheme performs best.
To validate the effectiveness of training pose-dependent skinning, we replace the SkinNet by a network predicting per-vertex displacement, which generalizes worse than SkinNet.
We train our model with sparse cameras (10 views), and the performance is still satisfying.

%
%
\section{Discussion and Conclusion} \label{sec:conclusion}
\textbf{Limitations.} 
Although we propose the first method for learning pose-dependent skinning solely from multi-view video, our method still has some limitations, which opens up interesting directions for future work.
Currently, we query the SkinNet for every vertex in canonical space, which may be inefficient when many points are sampled.
Thus, future work could involve exploring more efficient architectures like hashgrids~\cite{mueller2022instant}.
While hand and body can be faithfully animated, there is no control over the facial expression. 
Thus, extending our automated character creation pipeline with some parametric model for facial expressions might be a promising direction.
Also, a joint consideration of rigging, skinning, and pose tracking may further improve the quality of the 3D animation-ready characters.
\par
\textbf{Conclusion.}
We proposed the first method for creating a fully rigged and skinned character solely from multi-view video without any manual processing.
To this end, we first acquire a canonical template using an implicit surface reconstruction, automated initial skinning and mesh parsing.
Most importantly, we learn a pose-dependent skinning field from a multi-view video, which greatly reduces artifacts.
We show how such a skinning field can be learned in a weakly supervised manner by an appearance model and differentiable rendering.
Our work demonstrates that animatable character creation can be fully automated while maintaining high-quality geometry, rigging, and skinning.


{\small
\bibliographystyle{ieee_fullname}
\bibliography{mybib}

\begin{thebibliography}{10}\itemsep=-1pt

\bibitem{tensorflow}
Mart\'{\i}n Abadi, Ashish Agarwal, Paul Barham, Eugene Brevdo, Zhifeng Chen,
  Craig Citro, Greg~S. Corrado, Andy Davis, Jeffrey Dean, Matthieu Devin,
  Sanjay Ghemawat, Ian Goodfellow, Andrew Harp, Geoffrey Irving, Michael Isard,
  Yangqing Jia, Rafal Jozefowicz, Lukasz Kaiser, Manjunath Kudlur, Josh
  Levenberg, Dan Man\'{e}, Rajat Monga, Sherry Moore, Derek Murray, Chris Olah,
  Mike Schuster, Jonathon Shlens, Benoit Steiner, Ilya Sutskever, Kunal Talwar,
  Paul Tucker, Vincent Vanhoucke, Vijay Vasudevan, Fernanda Vi\'{e}gas, Oriol
  Vinyals, Pete Warden, Martin Wattenberg, Martin Wicke, Yuan Yu, and Xiaoqiang
  Zheng.
\newblock {TensorFlow}: Large-scale machine learning on heterogeneous systems,
  2015.
\newblock Software available from tensorflow.org.

\bibitem{agisoftphotoscan}
{Agisoft}.
\newblock {PhotoScan}.
\newblock \url{http://www.agisoft.com}, 2016.

\bibitem{rangescan}
Brett Allen, Brian Curless, and Zoran Popovi\'{c}.
\newblock Articulated body deformation from range scan data.
\newblock {\em ACM Trans. Graph.}, 21(3):612–619, jul 2002.

\bibitem{correlated}
Brett Allen, Brian Curless, Zoran Popovic, and Aaron Hertzmann.
\newblock {Learning a Correlated Model of Identity and Pose-Dependent Body
  Shape Variation for Real-Time Synthesis}.
\newblock In Marie-Paule Cani and James O'Brien, editors, {\em ACM SIGGRAPH /
  Eurographics Symposium on Computer Animation}. The Eurographics Association,
  2006.

\bibitem{anguelov05}
Dragomir Anguelov, Praveen Srinivasan, Daphne Koller, Sebastian Thrun, Jim
  Rodgers, and James Davis.
\newblock {SCAPE: Shape Completion and Animation of People}.
\newblock {\em ACM Transactions on Graphics}, 24(3):408--416, 2005.

\bibitem{splineskinning}
Seungbae Bang and Sung-Hee Lee.
\newblock Spline interface for intuitive skinning weight editing.
\newblock {\em ACM Trans. Graph.}, 37(5), sep 2018.

\bibitem{baran07}
Ilya Baran and Jovan Popovi\'{c}.
\newblock Automatic rigging and animation of 3d characters.
\newblock {\em ACM Trans. Graph.}, 26(3), July 2007.

\bibitem{chen2021snarf}
Xu Chen, Yufeng Zheng, Michael~J Black, Otmar Hilliges, and Andreas Geiger.
\newblock Snarf: Differentiable forward skinning for animating non-rigid neural
  implicit shapes.
\newblock In {\em International Conference on Computer Vision (ICCV)}, 2021.

\bibitem{voxelbinding}
Olivier Dionne and Martin de Lasa.
\newblock Geodesic voxel binding for production character meshes.
\newblock In {\em Proceedings of the 12th ACM SIGGRAPH/Eurographics Symposium
  on Computer Animation}, SCA '13, page 173–180, New York, NY, USA, 2013.
  Association for Computing Machinery.

\bibitem{geodesicbinding}
Olivier Dionne and Martin de Lasa.
\newblock Geodesic binding for degenerate character geometry using sparse
  voxelization.
\newblock {\em IEEE Transactions on Visualization and Computer Graphics},
  20(10):1367--1378, 2014.

\bibitem{arcspline}
Sven Forstmann, Jun Ohya, Artus Krohn-Grimberghe, and Ryan McDougall.
\newblock Deformation styles for spline-based skeletal animation.
\newblock In {\em Proceedings of the 2007 ACM SIGGRAPH/Eurographics Symposium
  on Computer Animation}, SCA '07, page 141–150, Goslar, DEU, 2007.
  Eurographics Association.

\bibitem{icml2020_2086}
Amos Gropp, Lior Yariv, Niv Haim, Matan Atzmon, and Yaron Lipman.
\newblock Implicit geometric regularization for learning shapes.
\newblock In {\em Proceedings of Machine Learning and Systems 2020}, pages
  3569--3579. 2020.

\bibitem{habermann21}
Marc Habermann, Lingjie Liu, Weipeng Xu, Michael Zollhoefer, Gerard Pons-Moll,
  and Christian Theobalt.
\newblock Real-time deep dynamic characters.
\newblock {\em ACM Trans. Graph.}, 40(4), jul 2021.

\bibitem{habermann20}
Marc Habermann, Weipeng Xu, Michael Zollhoefer, Gerard Pons-Moll, and Christian
  Theobalt.
\newblock Deepcap: Monocular human performance capture using weak supervision.
\newblock {\em Proceedings of the Conference on Computer Vision and Pattern
  Recognition (CVPR)}, 1:1, 2020.

\bibitem{habermann19}
Marc Habermann, Weipeng Xu, Michael Zollh\"{o}fer, Gerard Pons-Moll, and
  Christian Theobalt.
\newblock Livecap: Real-time human performance capture from monocular video.
\newblock {\em ACM Transactions on Graphics (TOG)}, 38(2):14:1--14:17, 2019.

\bibitem{Hyun2005SweepbasedHD}
Dae-Eun Hyun, Seung-Hyun Yoon, Jung-Woo Chang, Joon-Kyung Seong, Myung-Soo Kim,
  and Bert J{\"u}ttler.
\newblock Sweep-based human deformation.
\newblock {\em The Visual Computer}, 21:542--550, 2005.

\bibitem{boundedbiharmonic}
Alec Jacobson, Ilya Baran, Jovan Popovi\'{c}, and Olga Sorkine-Hornung.
\newblock Bounded biharmonic weights for real-time deformation.
\newblock {\em Commun. ACM}, 57(4):99–106, apr 2014.

\bibitem{skinmeshani}
Doug~L. James and Christopher~D. Twigg.
\newblock Skinning mesh animations.
\newblock {\em ACM Trans. Graph.}, 24(3):399–407, jul 2005.

\bibitem{jiang2022hifecap}
Yue Jiang, Marc Habermann, Vladislav Golyanik, and Christian Theobalt.
\newblock Hifecap: Monocular high-fidelity and expressive capture of human
  performances.
\newblock In {\em BMVC}, 2022.

\bibitem{kavan07}
Ladislav Kavan, Steven Collins, Ji{\v{r}}{\'\i} {\v{Z}}{\'a}ra, and Carol
  O'Sullivan.
\newblock Skinning with dual quaternions.
\newblock In {\em Proceedings of the 2007 symposium on Interactive 3D graphics
  and games}, pages 39--46. ACM, 2007.

\bibitem{elasticity}
Ladislav Kavan and Olga Sorkine.
\newblock Elasticity-inspired deformers for character articulation.
\newblock {\em ACM Trans. Graph.}, 31(6), nov 2012.

\bibitem{sphericalskinning}
Ladislav Kavan and Ji\v{r}\'{\i} \v{Z}\'{a}ra.
\newblock Spherical blend skinning: A real-time deformation of articulated
  models.
\newblock In {\em Proceedings of the 2005 Symposium on Interactive 3D Graphics
  and Games}, I3D '05, page 9–16, New York, NY, USA, 2005. Association for
  Computing Machinery.

\bibitem{kingma14}
Diederik Kingma and Jimmy Ba.
\newblock Adam: A method for stochastic optimization.
\newblock {\em International Conference on Learning Representations}, 12 2014.

\bibitem{robustaccurate}
Binh~Huy Le and Zhigang Deng.
\newblock Robust and accurate skeletal rigging from mesh sequences.
\newblock {\em ACM Trans. Graph.}, 33(4), jul 2014.

\bibitem{centersofrot}
Binh~Huy Le and Jessica~K. Hodgins.
\newblock Real-time skeletal skinning with optimized centers of rotation.
\newblock {\em ACM Trans. Graph.}, 35(4), jul 2016.

\bibitem{lbs}
J.~P. Lewis, Matt Cordner, and Nickson Fong.
\newblock Pose space deformation: A unified approach to shape interpolation and
  skeleton-driven deformation.
\newblock In {\em Proceedings of the 27th Annual Conference on Computer
  Graphics and Interactive Techniques}, SIGGRAPH '00, page 165–172, USA,
  2000. ACM Press/Addison-Wesley Publishing Co.

\bibitem{li2021learning}
Peizhuo Li, Kfir Aberman, Rana Hanocka, Libin Liu, Olga Sorkine-Hornung, and
  Baoquan Chen.
\newblock Learning skeletal articulations with neural blend shapes.
\newblock {\em ACM Transactions on Graphics (TOG)}, 40(4):1, 2021.

\bibitem{li2020self}
Peike Li, Yunqiu Xu, Yunchao Wei, and Yi Yang.
\newblock Self-correction for human parsing.
\newblock {\em IEEE Transactions on Pattern Analysis and Machine Intelligence},
  2020.

\bibitem{li2022tava}
Ruilong Li, Julian Tanke, Minh Vo, Michael Zollhofer, Jurgen Gall, Angjoo
  Kanazawa, and Christoph Lassner.
\newblock Tava: Template-free animatable volumetric actors.
\newblock 2022.

\bibitem{li2020deep}
Yue Li, Marc Habermann, Bernhard Thomaszewski, Stelian Coros, Thabo Beeler, and
  Christian Theobalt.
\newblock Deep physics-aware inference of cloth deformation for monocular human
  performance capture.
\newblock 2020.

\bibitem{liao2022pose}
Zhouyingcheng Liao, Jimei Yang, Jun Saito, Gerard Pons-Moll, and Yang Zhou.
\newblock Skeleton-free pose transfer for stylized 3d characters.
\newblock In {\em European Conference on Computer Vision ({ECCV})}. {Springer},
  October 2022.

\bibitem{liu2021neural}
Lingjie Liu, Marc Habermann, Viktor Rudnev, Kripasindhu Sarkar, Jiatao Gu, and
  Christian Theobalt.
\newblock Neural actor: Neural free-view synthesis of human actors with pose
  control.
\newblock {\em ACM Trans. Graph.}, 40(6), dec 2021.

\bibitem{neuroskinning}
Lijuan Liu, Youyi Zheng, Di Tang, Yi Yuan, Changjie Fan, and Kun Zhou.
\newblock Neuroskinning: Automatic skin binding for production characters with
  deep graph networks.
\newblock {\em ACM Trans. Graph.}, 38(4), jul 2019.

\bibitem{loper15}
Matthew Loper, Naureen Mahmood, Javier Romero, Gerard Pons-Moll, and Michael~J.
  Black.
\newblock {SMPL}: A skinned multi-person linear model.
\newblock {\em ACM Trans. Graphics (Proc. SIGGRAPH Asia)}, 34(6):248:1--248:16,
  Oct. 2015.

\bibitem{marchingcubes}
William Lorensen and Harvey Cline.
\newblock Marching cubes: A high resolution 3d surface construction algorithm.
\newblock {\em ACM SIGGRAPH Computer Graphics}, 21:163--, 08 1987.

\bibitem{Magnenat-Thalmann1988:4}
N. Magnenat-Thalmann, A. Laperri{`{e}}re, and D. Thalmann.
\newblock Joint-dependent local deformations for hand animation and object
  grasping.
\newblock In {\em Proceedings of Graphics Interface '88}, GI '88, pages 26--33.
  Canadian Man-Computer Communications Society, 1988.

\bibitem{mildenhall2020nerf}
Ben Mildenhall, Pratul~P. Srinivasan, Matthew Tancik, Jonathan~T. Barron, Ravi
  Ramamoorthi, and Ren Ng.
\newblock Nerf: Representing scenes as neural radiance fields for view
  synthesis.
\newblock In Andrea Vedaldi, Horst Bischof, Thomas Brox, and Jan-Michael Frahm,
  editors, {\em Computer Vision -- ECCV 2020}, pages 405--421, Cham, 2020.
  Springer International Publishing.

\bibitem{building}
Alex Mohr and Michael Gleicher.
\newblock Building efficient, accurate character skins from examples.
\newblock {\em ACM Trans. Graph.}, 22(3):562–568, jul 2003.

\bibitem{skinningnet}
A. Mosella-Montoro and J. Ruiz-Hidalgo.
\newblock Skinningnet: Two-stream graph convolutional neural network for
  skinning prediction of synthetic characters.
\newblock In {\em 2022 IEEE/CVF Conference on Computer Vision and Pattern
  Recognition (CVPR)}, pages 18572--18581, Los Alamitos, CA, USA, jun 2022.
  IEEE Computer Society.

\bibitem{bonehelpers}
Tomohiko Mukai and Shigeru Kuriyama.
\newblock Efficient dynamic skinning with low-rank helper bone controllers.
\newblock {\em ACM Trans. Graph.}, 35(4), jul 2016.

\bibitem{mueller2022instant}
Thomas M\"uller, Alex Evans, Christoph Schied, and Alexander Keller.
\newblock Instant neural graphics primitives with a multiresolution hash
  encoding.
\newblock {\em ACM Trans. Graph.}, 41(4):102:1--102:15, July 2022.

\bibitem{Park_2019_CVPR}
Jeong~Joon Park, Peter Florence, Julian Straub, Richard Newcombe, and Steven
  Lovegrove.
\newblock Deepsdf: Learning continuous signed distance functions for shape
  representation.
\newblock In {\em Proceedings of the IEEE/CVF Conference on Computer Vision and
  Pattern Recognition (CVPR)}, June 2019.

\bibitem{capturingskin}
Sang~Il Park and Jessica~K. Hodgins.
\newblock Capturing and animating skin deformation in human motion.
\newblock {\em ACM Trans. Graph.}, 25(3):881–889, jul 2006.

\bibitem{peng2021animatable}
Sida Peng, Junting Dong, Qianqian Wang, Shangzhan Zhang, Qing Shuai, Hujun Bao,
  and Xiaowei Zhou.
\newblock Animatable neural radiance fields for human body modeling.
\newblock {\em ICCV}, 2021.

\bibitem{peng2020neural}
Sida Peng, Yuanqing Zhang, Yinghao Xu, Qianqian Wang, Qing Shuai, Hujun Bao,
  and Xiaowei Zhou.
\newblock Neural body: Implicit neural representations with structured latent
  codes for novel view synthesis of dynamic humans.
\newblock {\em CVPR}, 1(1):9054--9063, 2021.

\bibitem{scanimate}
Shunsuke Saito, Jinlong Yang, Qianli Ma, and Michael~J. Black.
\newblock {SCANimate}: Weakly supervised learning of skinned clothed avatar
  networks.
\newblock In {\em Proceedings IEEE/CVF Conf.~on Computer Vision and Pattern
  Recognition (CVPR)}, June 2021.

\bibitem{salimans2016weight}
Tim Salimans and Durk~P Kingma.
\newblock Weight normalization: A simple reparameterization to accelerate
  training of deep neural networks.
\newblock {\em Advances in neural information processing systems}, 29, 2016.

\bibitem{BMSengupta20}
Soumyadip Sengupta, Vivek Jayaram, Brian Curless, Steve Seitz, and Ira
  Kemelmacher-Shlizerman.
\newblock Background matting: The world is your green screen.
\newblock In {\em Computer Vision and Pattern Regognition (CVPR)}, 2020.

\bibitem{Sloan01}
Peter-Pike~J. Sloan, Charles~F. Rose, and Michael~F. Cohen.
\newblock Shape by example.
\newblock In {\em Proceedings of the 2001 Symposium on Interactive 3D
  Graphics}, I3D '01, pages 135--143, New York, NY, USA, 2001. Association for
  Computing Machinery.

\bibitem{captury}
{TheCaptury}.
\newblock {The Captury}.
\newblock \url{http://www.thecaptury.com/}, 2020.

\bibitem{2020SciPy-NMeth}
Pauli Virtanen, Ralf Gommers, Travis~E. Oliphant, Matt Haberland, Tyler Reddy,
  David Cournapeau, Evgeni Burovski, Pearu Peterson, Warren Weckesser, Jonathan
  Bright, St{\'e}fan~J. {van der Walt}, Matthew Brett, Joshua Wilson, K.~Jarrod
  Millman, Nikolay Mayorov, Andrew R.~J. Nelson, Eric Jones, Robert Kern, Eric
  Larson, C~J Carey, {\.I}lhan Polat, Yu Feng, Eric~W. Moore, Jake
  {VanderPlas}, Denis Laxalde, Josef Perktold, Robert Cimrman, Ian Henriksen,
  E.~A. Quintero, Charles~R. Harris, Anne~M. Archibald, Ant{\^o}nio~H. Ribeiro,
  Fabian Pedregosa, Paul {van Mulbregt}, and {SciPy 1.0 Contributors}.
\newblock {{SciPy} 1.0: Fundamental Algorithms for Scientific Computing in
  Python}.
\newblock {\em Nature Methods}, 17:261--272, 2020.

\bibitem{wang2021neus}
Peng Wang, Lingjie Liu, Yuan Liu, Christian Theobalt, Taku Komura, and Wenping
  Wang.
\newblock Neus: Learning neural implicit surfaces by volume rendering for
  multi-view reconstruction.
\newblock {\em NeurIPS}, 2021.

\bibitem{arah}
Shaofei Wang, Katja Schwarz, Andreas Geiger, and Siyu Tang.
\newblock Arah: Animatable volume rendering of articulated human sdfs.
\newblock In {\em Computer Vision – ECCV 2022: 17th European Conference, Tel
  Aviv, Israel, October 23–27, 2022, Proceedings, Part XXXII}, page 1–19,
  Berlin, Heidelberg, 2022. Springer-Verlag.

\bibitem{multiweight}
Xiaohuan~Corina Wang and Cary Phillips.
\newblock Multi-weight enveloping: Least-squares approximation techniques for
  skin animation.
\newblock In {\em Proceedings of the 2002 ACM SIGGRAPH/Eurographics Symposium
  on Computer Animation}, SCA '02, page 129–138, New York, NY, USA, 2002.
  Association for Computing Machinery.

\bibitem{boneglow}
Rich Wareham and Joan Lasenby.
\newblock Bone glow: An improved method for the assignment of weights for mesh
  deformation.
\newblock In Francisco~J. Perales and Robert~B. Fisher, editors, {\em
  Articulated Motion and Deformable Objects}, pages 63--71, Berlin, Heidelberg,
  2008. Springer Berlin Heidelberg.

\bibitem{RigNet}
Zhan Xu, Yang Zhou, Evangelos Kalogerakis, Chris Landreth, and Karan Singh.
\newblock Rignet: Neural rigging for articulated characters.
\newblock {\em ACM Trans. on Graphics}, 39, 2020.

\bibitem{yang2022object}
Ji Yang, Xinxin Zuo, Sen Wang, Zhenbo Yu, Xingyu Li, Bingbing Ni, Minglun Gong,
  and Li Cheng.
\newblock Object wake-up: 3d object rigging from a single image.
\newblock In {\em Computer Vision--ECCV 2022: 17th European Conference, Tel
  Aviv, Israel, October 23--27, 2022, Proceedings, Part II}, pages 311--327.
  Springer, 2022.

\bibitem{Yang2006CurveSS}
Xiaosong Yang, Arun Somasekharan, and Jian~Jun Zhang.
\newblock Curve skeleton skinning for human and creature characters.
\newblock {\em Computer Animation and Virtual Worlds}, 17, 2006.

\end{thebibliography}
}

\appendix

\section{Appendix}
In the following, we show quantitative and qualitative comparisons against ARAH~\cite{arah} and TAVA~\cite{li2022tava} (Sec.~\ref{supp_sec:comparison}). 
Then, we provide more details concerning our method (Sec.~\ref{supp_sec:parsing}-\ref{supp_sec:training}).
We also provide additional information about how the competing approaches are trained and evaluated (Sec.~\ref{supp_sec:competing}).
Moreover, we provide more qualitative results that compare our design choice of using NeuS~\cite{wang2021neus} for template generation with classical multi-view stereo reconstruction~\cite{agisoftphotoscan} (Sec.~\ref{supp_sec:mvs}).
Lastly, we demonstrate that our method can pose template meshes of varying resolution without re-training (Sec.~\ref{supp_sec:multires}).

\section{Additional Comparisons} \label{supp_sec:comparison}
%
%
%
\begin{table}
\begin{center}
\begin{tabular}{|c|c|c|c|}
    \hline
    \multicolumn{4}{|c|}{\textbf{\textit{Quantitative Geometry Comparison}}} \\
    \hline
    \textbf{Subject} & \multicolumn{3}{|c|}{D2} \\
    \hline
    \textbf{Method}  & \textbf{Chamfer}$\downarrow$  & \textbf{M2S}$\downarrow$ & \textbf{S2M}$\downarrow$  \\
    \hline
    Initial weights~\cite{baran07} 	                             &  3.760               &  2.162         &   1.599   \\
    ARAH~\cite{arah}                                  &   12.985               &  7.625         &   5.360   \\
    TAVA~\cite{li2022tava} 	                            &   5.369               &  3.017         &   2.351   \\
    \textbf{Ours} 	                        &   \textbf{3.034}               &  \textbf{1.746}         &   \textbf{1.288}  \\
    \hline
    \end{tabular}
    \end{center}
    \caption
    {
    We further compare our method to ARAH~\cite{arah} and TAVA~\cite{li2022tava} on D2.
    As a reference, we also show the results with initial skinning weights obtained from Pinocchio~\cite{baran07}.
    Our method clearly outperforms both approaches and the baseline as TAVA~\cite{li2022tava} leverages a density-based surface representation that usually fails to model high quality geometry, and both cannot scale to the more difficult DynaCap dataset, which contains significantly more frames and pose variations as the dataset used in their work.
    }
	\label{tab:supp_comparison}	
\end{table}
%
%

%
%
%
\begin{figure*}
	\centering
	\includegraphics[width=0.95\textwidth]{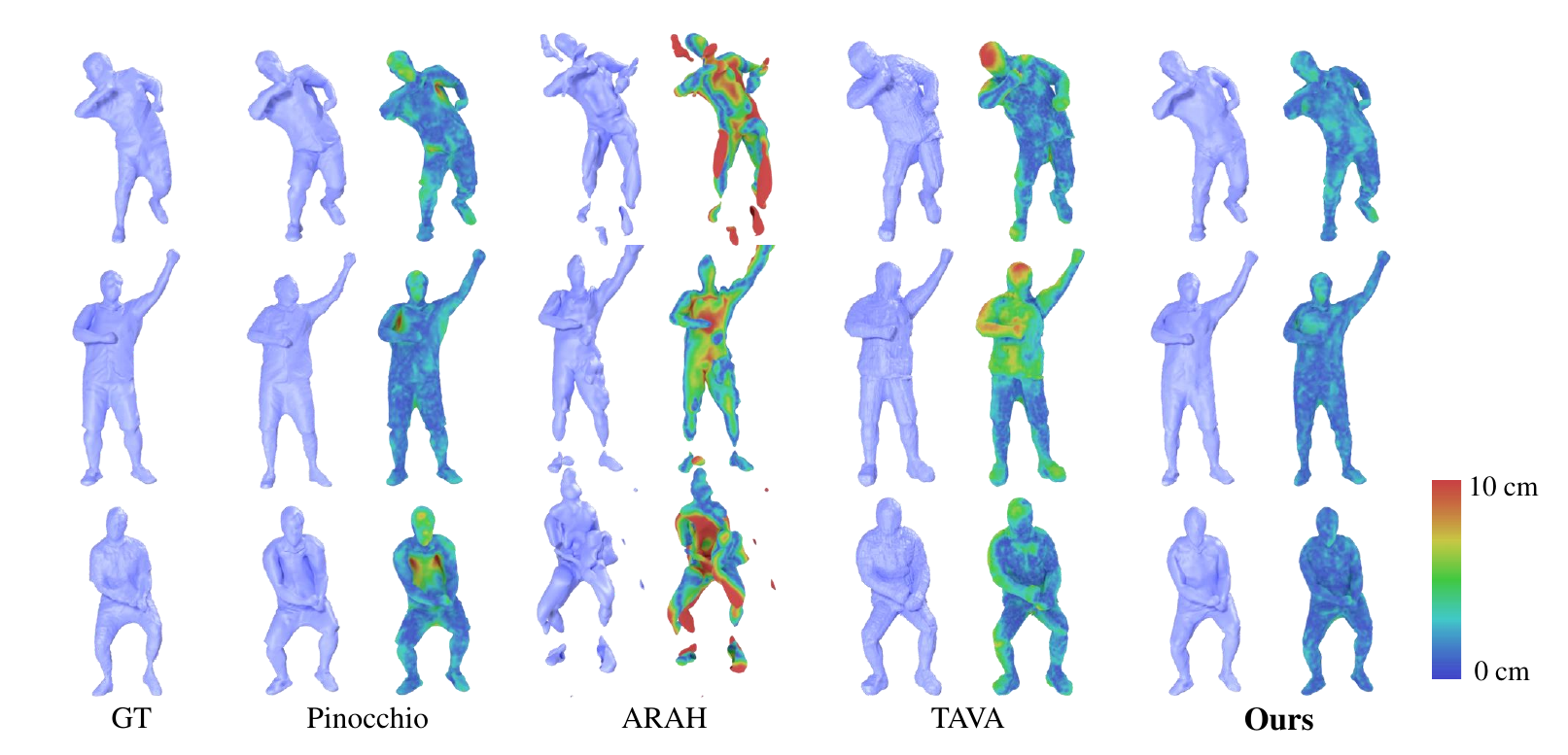}
	\caption
	{
	    Qualitative comparison.
	    For each method, we visualize the recovered posed geometry as well as the per-vertex error map when comparing the ground truth in terms of M2S.
	    Note that our method consistently shows the lowest error and also has the least visual artifacts.
	}
	\label{fig:supp_comparison}
\end{figure*}
%
%
%
We compare VINECS against ARAH~\cite{arah} and TAVA~\cite{li2022tava}, which are animatable human volume rendering methods, on the subject D2.
We provide the quantitative results in Tab.~\ref{tab:supp_comparison} and the qualitative results in Fig.~\ref{fig:supp_comparison}.
The evaluation protocol is the same one as  described in the main manuscript.
Their reconstruction error is significantly higher than the one of VINECS. 
Interestingly, their error is also higher compared to using initial skinning weights.
We found two major reasons for this:
1) The DynaCap dataset \cite{habermann21} is significantly more challenging than the datasets used for evaluation in their works, and their methods seem to not scale well to more complex conditions (motions, lighting).
2) For TAVA, we found that they use a density-based representation for which it is hard to extract accurate geometry.
In contrast, our approach can scale well and achieves a higher accuracy.
\par
We would also like to highlight that their setting and goal is different from ours.
Their goal is to obtain an implicit animatable human model for volume rendering, whilst ours is to obtain pose-dependent skinning for explicit human animation.

%
%
\section{Human Parsing Labels} 
\label{supp_sec:parsing}
To obtain the per-vertex human parsing labels, we first apply a 2D human parsing method~\cite{li2020self} pre-trained on the LIP dataset on renderings from multiple views.
More specifically, we render the template mesh for one frame, animated by the initial skinning weights colored by the texture obtained by NeuS~\cite{wang2021neus}. We apply ambient lighting for the rendering.
We found for certain views, especially views from the top, the human parsing method often failed. Thus, we discarded such views.
Then, we run the human parsing method for the rendering of the remaining views. 
For each view, we can obtain a label for each vertex by finding the label of the nearest 2D pixel from its projection, and we perform max-voting to obtain the final label.
Note that if the label with the most votes is the background, we select the label with the second most votes instead.
After that, we iteratively run a mode filter within one-ring neighborhood until no vertex is labeled as the background.
Originally, the method of Li \textit{et  al.}~\cite{li2020self} predicts $20$ classes. 
We merge these classes and only keep two classes for our training, \textit{i.e., }the skin and the clothes. 
%
%
%
\section{Network Architectures} \label{supp_sec:skinning}
All the neural networks used in this paper, namely SkinNet, AlbedoNet, and ReflectanceNet, are based on the coordinate-based multi-layer perceptrons (MLP) 
sharing the same set of hyper-parameters. 
The network contains five layers with $256, 256, 128, 256$ and $256$ neurons in them, respectively. 
There is a skip connection from the input to the third layer.
Inspired by~\cite{icml2020_2086}, we use SoftPlus as the activation function. 
When the query point is fed into the network, we compute its positional encoding~\cite{mildenhall2020nerf} and concatenate it with the 3D coordinate. 
In addition, we re-parameterize the network weights using the weight normalization~\cite{salimans2016weight}.

For SkinNet, a SoftMax layer is applied on the output of the MLP, along the dimension of joints, to ensure that the output skinning weights satisfy the partition of unity. For ReflectanceNet and for AlbedoNet, there is no processing of the output during training, while during inference, we clip the values so that they are in the range $[0; 1]$.
As for the output of ReflectanceNet, we compute its exponential as the final scalar multiplier.

%
%
%
\section{Silhouette Loss} \label{supp_sec:silhouette}
It is a bidirectional loss, which aims to align the projected boundary of the mesh to the foreground mask $\mask$:
%
\begin{align}
    \loss_{silh} = \sum_{c=1}^{C} ( & \sum_{i \in B_c} \parallel d_c(\pi_c(\x_i)) \parallel^2 + \nonumber \\
    + & \sum_{\p \in \{ \mathbf{u} \in \R^2 \| d_c(\mathbf{u}) = 0 \}} \parallel \pi_c(\x_\p) - \p \parallel^2  )
    .
\end{align}
%
Specifically, the first term pushes boundary vertices $B_c$ to the boundary of the foreground mask, where the distance transform value $d_c$ equals zero. 
In the second term, we minimize the distance between every pixel $\p$ on the zero contour of the distance transform map and its closest projected vertex $\x_\p$. 
%
%
%
\section{Training Details} \label{supp_sec:training}
We implement our method using TensorFlow~\cite{tensorflow}. 
All experiments are performed on a single NVIDIA A40 GPU (48GB). Our training consists of four stages. 
During the first stage, we train SkinNet alone without the rendering loss for $50000$ iterations. 
Next, we train AlbedoNet for $5000$ iterations and ReflectanceNet for $5000$ iterations. 
Lastly, SkinNet is refined with the pre-trained appearance field for $20000$ iterations. The whole training takes around $18$ hours.

For all training stages, the network weights are optimized using Adam~\cite{kingma14}. We clip the gradient values to $[-1, 1]$. The learning rate is $0.001$ and the batch size is $4$.
%
%
\section{Competing Methods} \label{supp_sec:competing}
\textbf{Pinocchio~\cite{baran07}.}
We re-implement Pinocchio using Python. Since in our work, the skeleton is obtained from~\cite{captury}, we only use its skin attachment part to compute the skinning weights based on the skeleton. We use the library SciPy~\cite{2020SciPy-NMeth} to solve the sparse linear system for the heat equilibrium equation in Pinocchio.
%

%
%
\par 
\textbf{SCANimate~\cite{scanimate}.}
SCANimate requires the registered SMPL~\cite{loper15} pose and shape parameters for all training scans.
As we already had the pose tracking of the training sequence, we can animate our template mesh using the initial skinning weights, and the animated meshes can roughly align the scans.
Thus, we manually label $30$ correspondence points between our template mesh and SMPL template and optimize SMPL parameters by fitting to the correspondence points on our animated template mesh.
With the paired scan and SMPL parameter, we train SCANimate for each of our characters.
%
%
\par 
\textbf{SCANimate*~\cite{scanimate}.}
Since the focus of our paper is to learn the skinning weights, instead of the pose-aware shape, we test SCANimate without the pose-aware shape.
More specifically, during inference, we input the canonical pose parameter to the pose-dependent geometry module to obtain a canonical shape, which we keep constant for all poses.
We use this canonical shape as the query for the forward skinning network to obtain the skinning weights, which then animates the canonical shape by LBS.

%
%
\par 
\textbf{RigNet~\cite{RigNet}.}
We ran RigNet pre-trained on ``ModelsResource-RigNetv1'' dataset on our template mesh to obtain the skinning weights. Similar to Pinocchio, we only use the skinning prediction module, which takes a mesh and the aligned skeleton and predicts the skinning weights.
\par
\textbf{ARAH~\cite{arah}.}
We train ARAH using the same hyper-parameter settings as in their public codes. It is trained for 1.5 days on 4 NVIDIA A40 GPUs, which is much more expensive than the training of VINECS. 
We extract the mesh from the SDF of ARAH using Marching Cube~\cite{marchingcubes} with the resolution of $256^3$.

\par
\textbf{TAVA~\cite{li2022tava}.}
We train TAVA following the same setting as in their public code. The whole training takes 30 hours on a single NVIDIA A40 GPU.
The mesh is extracted from the density field using Marching Cube with the resolution of $256^3$.
%
%
\section{MVS vs. NeuS} \label{supp_sec:mvs}
%
%
%
\begin{figure}
	\includegraphics[width=0.9\linewidth]{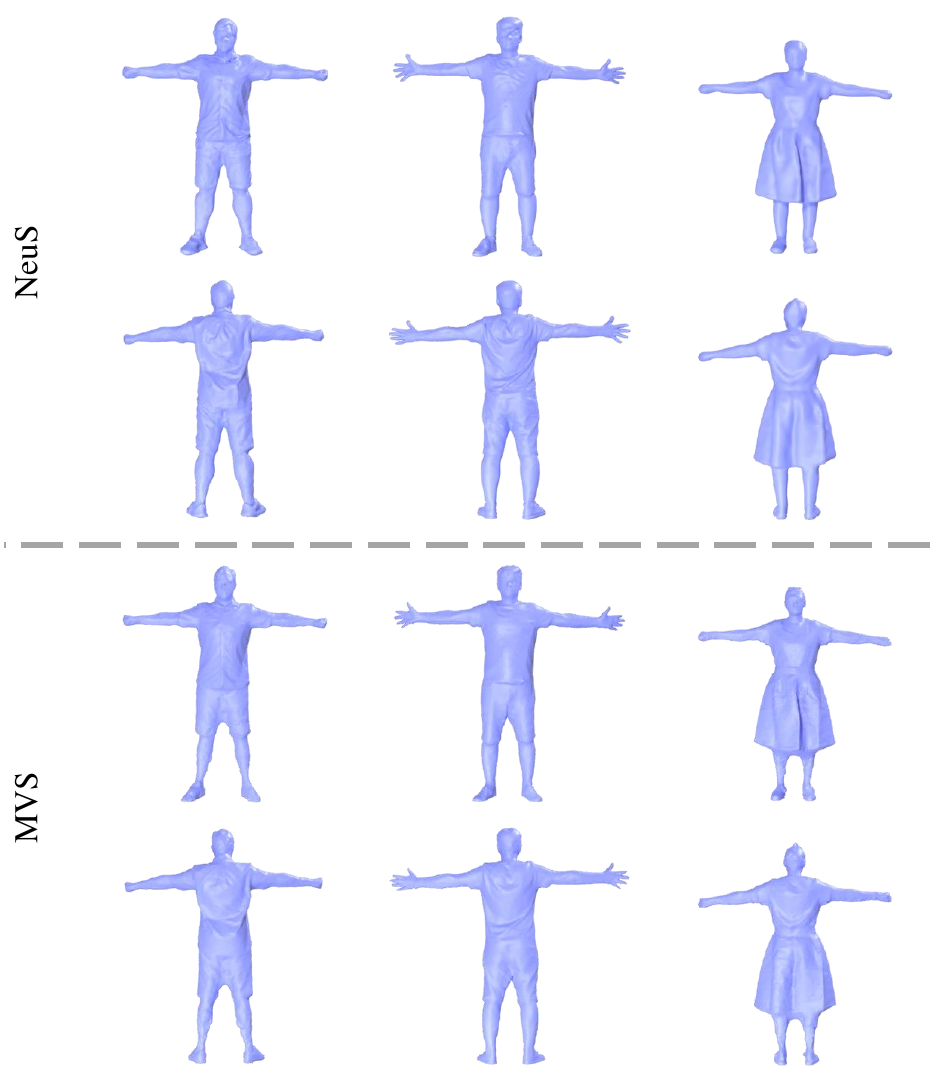}
	\caption
	{
	    NeuS~\cite{wang2021neus} (top) vs. Multi-view stereo reconstruction~\cite{agisoftphotoscan} (bottom).
	    In most cases, NeuS generates more high-frequency details (cloth wrinkles) while introducing less noise (e.g., around the calf).
	}
	\label{fig:mvs_vs_neus}
\end{figure}
%
%
%
We choose NeuS instead of classical multi-view stereo reconstruction to extract the template mesh because we found in most cases NeuS generates more high-frequency details while introducing less noise (see Fig.~\ref{fig:mvs_vs_neus}).
%
%
\section{Multi-resolution Results} \label{supp_sec:multires}
%
%
%
\begin{figure}
	\includegraphics[width=0.9\linewidth]{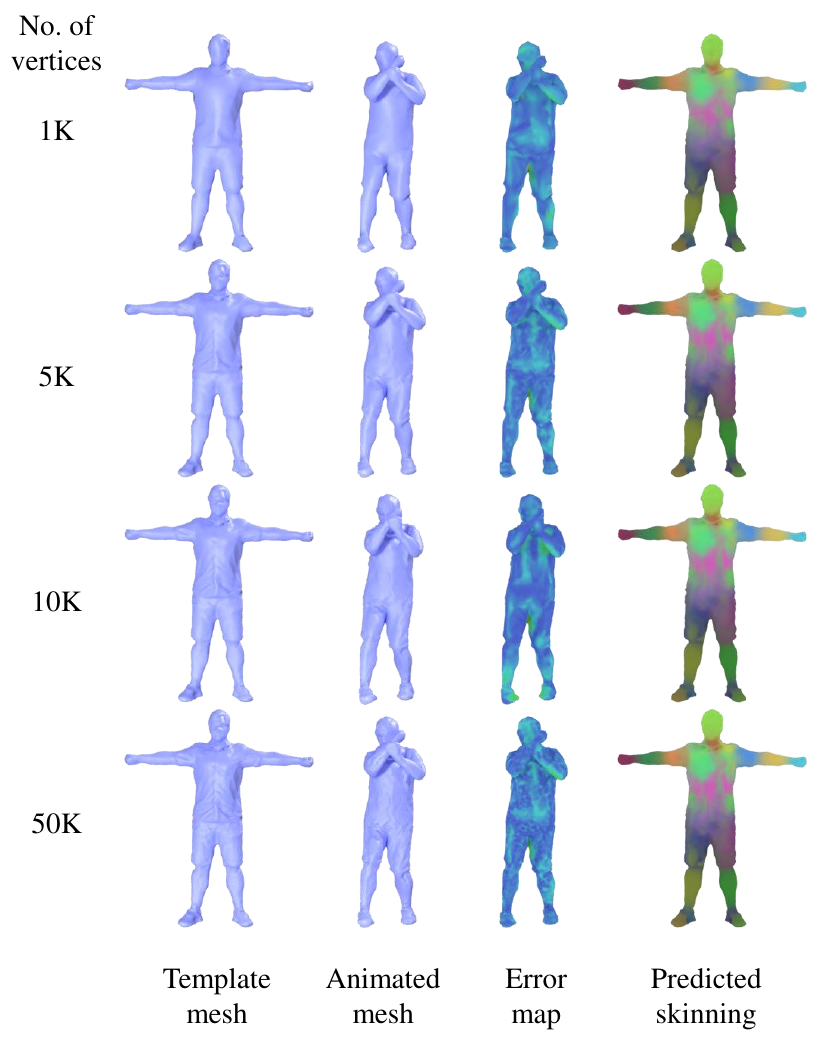}
	\caption
	{
	    Multi-resolution results of our method.
	    Even though trained on the fixed resolution (10K), our method generalizes well to other resolutions. Note that for different resolutions, the predicted skinning weights are very similar and the error always stays low.
	}
	\label{fig:multi-res}
\end{figure}
%
%
%
Our method supports multi-resolution character skinning because SkinNet is an implicit function and can take arbitrary 3D positions as input.
During training, we only input the vertices of the template mesh, which has a fixed resolution of around 10K vertices, to SkinNet, because our supervision requires an explicit mesh.
However, even though only trained on a fixed resolution, our method generalizes well to different resolutions.
We re-sample the original mesh generated by NeuS to different numbers of vertices (1K, 5K, 10K, 50K). Then, they are fed into the same pre-trained model and animated. All meshes deform naturally and have low 3D errors (Fig.~\ref{fig:multi-res}).
%


\end{document}